\documentclass{article}

\usepackage{arxiv}

\usepackage[utf8]{inputenc} 
\usepackage[T1]{fontenc}    
\usepackage{hyperref}       
\usepackage{url}            
\usepackage{booktabs}       
\usepackage{amsmath}
\usepackage{amsfonts}       
\usepackage{nicefrac}       
\usepackage{microtype}      
\usepackage{lipsum}
\usepackage{graphicx}
\graphicspath{ {./images/} }
\usepackage{bbm}
\usepackage{array}
\usepackage{hyperref}
\usepackage[ruled, vlined]{algorithm2e}
\usepackage{color}

\usepackage{colortbl}
\newcolumntype{P}[1]{>{\centering\arraybackslash}p{#1}}
\newcolumntype{R}[1]{>{\flushrifht\arraybackslash}p{#1}}

\definecolor{c_diag}{rgb}{0.85,0.89,0.953}
\definecolor{t_diag}{rgb}{0,0,0}
\definecolor{c_ndiag}{rgb}{.95,0.95,0.95}

\definecolor{c_hlight}{rgb}{.886,0.9411,0.851}
\definecolor{c_hdark}{rgb}{.663,0.82,0.557}
\definecolor{c_rest}{rgb}{.984,0.898,0.839}

\DeclareMathOperator*{\argmax}{\arg\!\max}

\title{XNB: Explainable Class--Specific Naïve--Bayes Classifier}

\author{
 Jes\'us S. Aguilar--Ruiz \\
  School of Engineering\\ Pablo de Olavide University\\ 
  ES-41013 Seville, Spain \\
  \texttt{aguilar@upo.es} \\
   \AND
  Cayetano Romero \\
  Minsait (Indra) \\
  Madrid, Spain \\
  \texttt{cromerov@minsait.com} \\
  \And
Andrea Cicconardi \\
Istituto Italiano di Tecnologia (IIT)\\ 
University of Genova, Italy \\
\texttt{andrea.cicconardi@iit.it} \\
}

\begin{document}
\maketitle
\begin{abstract}
In today's data--intensive landscape, where high--dimensional datasets are increasingly common, reducing the number of input features is essential to prevent overfitting and improve model accuracy. Despite numerous efforts to tackle dimensionality reduction, most approaches apply a universal set of features across all classes, potentially missing the unique characteristics of individual classes. This paper presents the Explainable Class--Specific Naïve Bayes (XNB) classifier, which introduces two critical innovations: 1) the use of Kernel Density Estimation to calculate posterior probabilities, allowing for a more accurate and flexible estimation process, and 2) the selection of class--specific feature subsets, ensuring that only the most relevant variables for each class are utilized. Extensive empirical analysis on high--dimensional genomic datasets shows that XNB matches the classification performance of traditional Naïve Bayes while drastically improving model interpretability. By isolating the most relevant features for each class, XNB not only reduces the feature set to a minimal, distinct subset for each class but also provides deeper insights into how the model makes predictions. This approach offers significant advantages in fields where both precision and explainability are critical.
\end{abstract}


\section{Introduction} \label{sec:introduction}
Predictive models are important because they provide forecasts that, on many occasions, help to make more reliable decisions in sensitive contexts. The usefulness and effectiveness of these models is widely proven in a multitude of application domains, receiving different names: predictive systems \cite{Gent2013}, classifier systems \cite{Ho1994} or diagnostic systems \cite{Swets1998}. 

Currently, there is a wide spectrum of techniques derived from machine learning that can be used to address the design of a predictive model. Most of them generate the model from the data (learning phase) and, once the quality of the model has been verified with some metric (evaluation phase) in a validation process (testing phase), it is applied to new unseen observations (deployment phase). The choice of model depends on several factors, including performance, quality, complexity, and explainability \cite{Barragan-Montero2022}.

Predictive techniques (classification or regression) differ considerably in their basis and properties and, therefore, may also generate uneven results. The choice of these techniques, individually or in combination, will depend on the characteristics of the problem, i.e., the specific requirements and objectives. Most of these techniques belong to the \textit{opaque} machine learning, which includes methods that do not provide knowledge model, and instead function as black boxes (e.g., neural networks), as opposed to the \textit{explainable} machine learning, which comprises techniques that generate an interpretable model capable of explaining (partially or totally) the knowledge extracted from data (e.g., decision trees). In short, \textit{explainable} means providing details about how the model works in order to better understand why a particular decision was made. 

In the broad field of Artificial Intelligence (AI) there has recently emerged a sudden and remarkable interest in understanding how the model makes decisions in the sense that humans can interpret the knowledge contained in the model \cite{Herrera2020}, named Explainable AI (XAI). Explainable models are increasingly in demand, especially in the field of biomedicine, where understanding can lead to progress, as unfortunately not all the knowledge resides in the model.

Classification addresses predictive tasks when the target variable is nominal (e.g., type of tumor). One of the simplest, fastest and most effective classification technique is naïve--Bayes (NB). So much so that it often becomes a comparative benchmark in the design of new predictive approaches. However, NB assumes that variables follow a normal distribution, which is a very optimistic assumption, as it is not met by all multimodal distributions, in addition to the unimodal non--Gaussian ones.  As the Gaussian model may not work well if the data does not fit the assumption, a non--parametric estimation of the data distribution could provide better fitness. This idea was already introduced in the Flexible Naïve Bayes approach (FNB)\cite{John1995}, which used Gaussian kernel functions to estimate the probability density function of the variables (independently) and incorporated them into the calculation of probabilities for the NB classifier. 

NB also includes another important assumption: variables are conditionally independent given the class. This assumption has enormous implications on the computational cost of the method, as variables can be examined independently.  However, when dealing with very high dimensional datasets (thousands of variables), as NB calculates the product of conditional probabilities, the posterior class probability tends quickly to zero. In this highly dimensional context, it would be very effective to determine the variables that are significant for each class, independently. 

This research addresses two important issues: a) How relevant is each variable for each class?; b) How to improve the posterior probability estimate, given the class? By posing answers to these two questions, it is found that a posterior probability estimation method can be used that allows using only a subset of relevant variables, which is different for each class, thus achieving a higher explanatory ability of the model while maintaining its classification performance. For example, to predict the tumor type of an unseen new case, among three possible tumor types, using several thousand variables, the method would provide a decision based on only 10 variables for type I, 4 variables for type II, and 7 variables for type III. Thus, not only could a better classification performance be achieved, but the model would also have an important explanatory power.

In summary, the new approach, named Explainable Class--Specific Naïve--Bayes (XNB) classifier includes two important features: a) the posterior probability is calculated by means of Kernel Density Estimation (KDE); b) the posterior probability for each class does not use all variables, but only those that are relevant for each specific class. From the point of view of the classification performance, the XNB classifier is comparable to NB classifier. However, the XNB classifier provides the subsets of relevant variables for each class, which contributes considerably to explaining how the predictive model is performing. In addition, the subsets of variables generated for each class are usually different and with remarkably small cardinality. 



The rest of the paper is organized as follows: Section \ref{sec:notation} describes the mathematical notation used hereafter; the fundamentals of NB are detailed in Section \ref{sec:nb}; the application of KDE for the calculation of probabilities, as well as its use in the FNB approach, is explained in Section \ref{sec:fnb}; the novel approach, named XNB, including a discussion on its class--specific focus and a detailed description of the method is provided in Section \ref{sec:XNB}; Section \ref{sec:experiments} presents a comparative analysis of NB and XNB in terms of classification performance and number of variables selected, carried out with eighteen very--high dimensional genomic datasets; finally, conclusions and future work are summarized in Section \ref{sec:conclusions}.

\section{Data notation}\label{sec:notation}

Let $D=(E,F,\upsilon,\omega)$ be a data set, where $E$ is the set of example (or instance) identifiers, $F$ is the set of feature (variable) identifiers, $\upsilon:E\times F \rightarrow \mathbb{R}$ is the function that assigns a real value to a pair $(e,f)$, where $e\in E$ and $f\in F$. Within the field of supervised learning, when $\omega:E \rightarrow \mathbb{L}$ assigns a class label $c$ to an example $e$, with $\mathbb{L}=\{c_1,\dots,c_k\}$, then it is a classification problem; otherwise, if $\mathbb{L}\subseteq\mathbb{R}$, then it is a regression problem. In absence of example identifiers (e.g. patient identifiers) or feature identifiers (e.g. gene names), it is convenient to use $E=\{e_1,\dots,e_n\}$ and $F=\{f_1,\dots,f_m\}$, respectively. Henceforth, $\upsilon(e_i,f_j)$ will represent a real value, and $\omega(e_i)$ will represent a class label (e.g. type of tumor). 

Both the function $\upsilon$ and the function $\omega$ are expressed in tabular form, so that $\upsilon$ is determined by a $n\times m$ matrix of real values, and $\omega$ as a vector of $n$ values from $\mathbb{L}$. In order to simplify the notation, a function is defined to extract the values of sample $e_i$: $input: E \rightarrow \mathbb{R}^m$, such that $input(e_i)=(\upsilon(e_i,f_1),\dots,\upsilon(e_i,f_m))$.

The main goal of any classification problem is to find a general function $\Omega: \mathbb{R}^m \rightarrow \mathbb{L}$, learned from $D$, such that $\Omega(input(e_i)) = \omega(e_i)$, $\forall e_i$, where $i=\{1,\dots,n\}$, or at least it maximizes the frequency of the equality (a popular --and biased due to its sensitivity to imbalance \cite{Aguilar2024a}-- measure of quality of $\Omega$ is the relative frequency of success, named \textit{accuracy}).

\section{Naïve--Bayes}\label{sec:nb}


The NB classifier predicts that a test sample $x=(x_1,$\dots,$x_m)$ with $m$ variables, belongs to the class $c_k \in \mathbb{L}$ if and only if $P(c_k\vert x)>P(c_{k'}\vert x)$, for $1\leq k'\leq |\mathbb{L}|$, with $k'\neq k$. The class $c_k$ for which $P(c_k\vert x)$ is maximized is called the \textit{maximum posteriori hypothesis} and can be calculated using the Bayes' theorem in Eq. \ref{eq:Bayes}.
\begin{equation} \label{eq:Bayes}
P(c_k\vert x)=\frac{P(x\vert c_k)P(c_k)}{P(x)}
\end{equation}

As $P(x)$ is constant for all classes, only $P(x\vert c_k)P(c_k)$ needs to be maximized, taking into account that $P(c_k)=\frac{|c_k|}{|D|}$, being $D$ the dataset, and $|c_k|$ the number of samples of class $c_k$ in $D$.

To compute $P(x \vert c_k)$ is extremely expensive, as it would mean to calculate conditional probabilities among variables. Instead, the \textit{naïve} assumption of class--conditional independence reduces dramatically the calculations to $P(x \vert c_k)=\prod_{j=1}^m P(x_j \vert c_k)$. Despite this strong assumption, which is usually not tested, it performs surprisingly well. 

In short, the process of classifying a new instance $x$ consists of choosing the class $c^*$ with the highest \textit{a posteriori} probability, as shown in Eq. \ref{eq:bayes_decision}. In NB that probability is obtained by estimating mean and variance for each numerical variable from data and using these two statistics in the Gaussian probability distribution expression.
\begin{equation}\label{eq:bayes_decision}   
c^*=\argmax \limits_{i \in \{1, \ldots, k\}} P(c_i) \displaystyle \prod_{j=1}^m P(x_j \vert c_i)
\end{equation}

Three critical assumptions emerge from the NB approach: a) the assumption of normality; b) the assumption of conditional independence of variables given the class; c) all the $m$ variables $x_j$ are equally important and therefore involved in the calculation of the $k$ probabilities for class $c_i$ (Eq. \ref{eq:bayes_decision}). All these issues are inherent to NB. The first one was already addressed by the FNB approach. The approach introduced in XNB addresses the third issue (and uses the approach introduced in FNB to deal with the assumption of normality), providing a method to reduce the feature space dimensionality that will be explained in detail in Sec. \ref{sec:XNB}.

Given the important of the first two assumptions, a verification exercise has been conducted to analyze the potential impact of the hypotheses of normality and conditional independence on the eighteen datasets included in the experimental analysis in Sec. \ref{sec:experiments}, and described in Table \ref{tab:datasets}. 


\begin{table}[t]
\caption{Description of CuMiDa datasets. First two columns identify the datasets: Cancer Type and GSE (Gene Expression Omnibus Series Accession number). \#s, \#v and \#c stand for number of samples, of variables and of classes, respectively. SW : Shapiro--Wilk's Test of normality; P: Pearson's Test of conditional independence. Last row (Mean) stands for the mean of \#s, of \#v, of \#c, SW and P, respectively.}
\label{tab:datasets}
\begin{center}
\begin{tabular}{|l|c||r|c|c||c|c|}
\hline
Type 		& GSE	& \#s 		& \#v 		& \#c 	& SW	& P		\\ \hline \hline
Bladder		& 31189	& 85			& 54,674		& 2		&	0.61	&	0.77	\\ \hline
Brain			& 15824	& 37			& 54,675		& 4		&	0.47	&	0.80	\\ \hline
Brain			& 50161	& 130		& 54,675		& 5		&	0.61	&	0.63	\\ \hline
Breast		& 07904	& 45			& 54,674 		& 3		&	0.38	&	0.65	\\ \hline
Breast		& 10797	& 66			& 22,277		& 3		&	0.84	&	0.39	\\ \hline
Breast		& 26304	& 115		& 33,636 		& 4		&	0.84	&	0.92	\\ \hline
Breast		& 42568	& 116		& 54,674		& 2		&	0.70	&	0.39	\\ \hline
Breast		& 45827	& 151		& 54,674 		& 6		&	0.68	&	0.38	\\ \hline
Colorectal		& 21510	& 147		& 54,674 		& 3		&	0.58	&	0.32	\\ \hline
Colorectal		& 77953	& 55			& 22,282		& 4		&	0.47	&	0.38	\\ \hline
Leukemia 		& 28497 	& 281 		& 22,282 		& 7 		&	0.76	&	0.28	\\ \hline
Lung			& 07670	& 51			& 22,283		& 2		&	0.49	&	0.82	\\ \hline
Ovary		& 06008	& 98			& 22,282		& 4		&	0.60	&	0.40	\\ \hline
Pancreatic	& 16515	& 51			& 54,675		& 2		&	0.48	&	0.75	\\ \hline
Prostate		& 11682	& 31			& 33,467		& 2		&	0.45	&	0.95	\\ \hline
Prostate		& 46602	& 49			& 54,675		& 2		&	0.31	&	0.47	\\ \hline
Renal		& 53757	& 143		& 54,675		& 2		&	0.56	&	0.43	\\ \hline
Throat		& 59102	& 42			& 32,703		& 2		&	0.41	&	0.73	\\ \hline \hline
Mean		&	--	& 94			& 42,109		& 3.3		&	0.57	&	0.58	\\ \hline  
\end{tabular}
\end{center}
\end{table}

\subsection{Normality assumption}

The Shapiro--Wilk normality test was performed with a significance level of 0.05, analyzing the eighteen datasets with a mean of 42,109 variables (see Table \ref{tab:datasets}: column SW). The SW test was applied to each variable in each dataset, and the percentage of variables rejecting the null hypothesis of normality (p-value$<$0.05) was recorded for each dataset (expressed in Table \ref{tab:datasets} as a ratio within [0,1]). The percentages of variables not following the normal distribution ranged from 31\% (GSE46602) to 84\% (GSE10797), highlighting the importance of the normality assumption. On average, 57\% of the variables rejected the null hypothesis, indicating that more than half of the variables do not follow a normal distribution. 

Despite the high percentage deviating from normality, the results provided by NB are generally very acceptable. However, the assumption of normality might become critical in datasets with tens of thousands of variables (e.g., genomic data \cite{Feltes2019}), or even millions of variables (e.g., transposable elements \cite{Wellinger2022}). 

A technique like KDE (see subsection \ref{subsec:kde}), which better adapts to non--normal distributions, could positively contribute to the classification process.

\subsection{Conditional independency assumption}

The assumption of conditional independence of variables given the class is often overlooked. However, this assumption can become significant in very high--dimensional datasets.

A conditional independence test was conducted to determine whether two continuous variables are independent given a third categorical variable (class). The Pearson test of conditional independence involves calculating the residuals of the continuous variables after regressing them on the categorical variable and then computing the Pearson correlation between these residuals. This partial correlation is then tested for significance to determine if the original variables are conditionally independent given the categorical variable. Partial correlation within the levels of the categorical variable was performed to identify all pairs of continuous variables that are conditionally dependent given the class. If the p-value is less than the significance level, the null hypothesis is rejected, indicating that the two continuous variables are not conditionally independent given the class. A very stringent significance level was set to ensure strong evidence against the null hypothesis $($p-value$<10^{-6})$. Additionally, it was required that the absolute value of the partial correlation coefficient be greater than 0.7. This ensured that the correlation was not only statistically significant but also practically significant (strong correlation). 

Table \ref{tab:datasets} (column P) shows the ratio of variables that are conditionally dependent on another variable given the class. The percentages range from 28\% (GSE28497) to 95\% (GSE11682), with an average of 58\%, revealing a strong conditional dependency among variables. The presence of dependency in 58\% of the variables suggests complex relationships. Many variables influence each other directly, beyond the influence of the class. This complexity might indicate underlying causal relationships not accounted for by the class. 

Significantly reducing the number of variables and selecting class--specific variables could positively contribute to mitigating conditional dependence in the calculation of posterior probabilities.






%





\section{Flexible Naive Bayes}\label{sec:fnb}

NB uses univariate Gaussians to calculate class--conditional marginal densities. However, the procedure can be generalized using one--dimensional kernel density estimates, which do not assume normal probability distribution.


Kernel Density Estimation (KDE) provides a more accurate way to calculate that probability since it does not assume any pre--defined probability distribution. Instead, it estimates from data the density distribution, from which probabilities can be easily calculated.

\subsection{Kernel Density Estimation} \label{subsec:kde}

Density estimation deals with the problem of estimating probability density functions based on some data sampled from the true unknown probability density function (\textit{pdf}).

In principle, a random variable can be fully described by its \textit{pdf}, which outlines the likelihood of a particular event happening when the random variable matches, or falls within a range close to, a specific value. However, selecting a \textit{pdf} model based on assumptions (a priori parametric choice) can be problematic in practice. This is because it may inaccurately represent the actual \textit{pdf}, for instance, by incorrectly assuming a normal distribution in situations where the data is actually multi--modal. A viable solution to this issue is to employ non--parametric \textit{pdf} estimators. These estimators allow us to derive the characteristics of the distribution directly from the data, sidestepping the pitfalls of making restrictive assumptions about the distribution's shape. Essentially, the key advantage of the non--parametric method is its flexibility, allowing for a more accurate representation of the distribution $f(x)$ without confining it to a predetermined form.

A well--known non--parametric estimator of the \textit{pdf} is the histogram. The histogram is easy to understand and to compute in one and higher dimensions. However, there are two important choices with high impact on the estimates: a) the starting point of each bin edge; b) and the bin size (bandwidth). Also, histograms estimators are usually not smooth, with discontinuous shapes (i.e., points within the same bin will provide the same estimation). This discrete nature of the \textit{pdf} is an intrinsic limit on resolution, which is highly depending on the selection of bandwidth and boundaries of the bins. Consequently, although it is a simple and fast technique, its usefulness is limited by its instability.

Kernel estimation of \textit{pdf} is a powerful technique, originated by Rosenblatt \cite{Rosenblatt1956} and Parzen \cite{Parzen1962}, that produces a smooth empirical \textit{pdf} based on the neighborhood of each individual sample from a given dataset. Let $\{x_1,\dots,x_n\}$ be an independent and identically distributed (iid) sample of $n$ observations taken from a population $X$ with unknown \textit{pdf}, $f(x)$. Kernel estimate of $f(x)$, $\hat{f}(x)$, assigns each $i$--th sample data point $x_i$ a function $K(\cdot)$ called a kernel function, which is non--negative and bounded for all $x$ ($0\leq K(x) < \infty$ $\forall x\in \mathbb{R}$).

Kernel functions are usually symmetric probability density functions that need a window width, named bandwidth, which determines the neighborhood size to be used in the estimation. The kernel density estimator at point $x$, also known as Parzen--Rosenblatt estimator, is an estimate of $f(x)$ based on the data, as defined in Eq. \ref{eq:kernel}.
\begin{equation}\label{eq:kernel}
\hat{f}_h(x)=\frac{1}{nh}\sum_{i=1}^nK\left(\frac{x-x_i}{h}\right)
\end{equation}

\noindent where $h>0$ is the bandwidth, $K(\cdot)$ is the kernel function and $n$ is the sample size.

%

The kernel $K(\cdot)$ is typically said to be of order $p \in \mathbb{N}$ if it satisfies Eq. \ref{eq:kde}
\begin{equation}\label{eq:kde}
\int_{-\infty}^{+\infty} x^j K(x) dx = 
\left\{
\begin{array}{ll}
1 & j=0  \\
0 & j=1,\dots,p-1  \\
\lambda & j=p  
\end{array}
\right.
\end{equation}
\noindent where the constant $\lambda \neq 0$ \cite{Hall1988}.

The most common kernels (uniform, Epanechnikov \cite{Epanechnikov1969}, biweight, triweight) are special cases of the beta polynomial family $B_s(x) = (1-x^2)^s$, where $s$ is a non--negative integer, and are defined in Eq. \ref{eq:beta}. These functions have a domain of $[-1,+1]$ and are orthogonal with respect to the beta distribution $B(\alpha,\beta)$, with parameters $\alpha=\beta=n+1$.
\begin{equation}\label{eq:beta}
K_s(x)=\frac{(2s+1)!!}{2^{s+1}s!}B_s(x) \mathbbm{1}(|x|\leq 1) 
\end{equation}

In fact, the most popular kernel, the Gaussian kernel defined in Eq. \ref{eq:gaussian}, is the limit of the beta polynomial function as the degree of the polynomial approaches infinity \cite{Duong2015}.
\begin{equation}\label{eq:gaussian}
K(x)=\frac{1}{\sqrt{2\pi}}e^{-\frac{1}{2}x^2}
\end{equation}

Usually, the symmetry assumption is imposed on the kernel $K(x)=K(-x)$, so that $p$ is necessarily even. As $\int K(x) dx =1$ (for $j=0$), when $K$ is non--negative, it is itself a probability density. Symmetric non--negative kernels are second--order kernels (e.g., the Gaussian kernel is of order 2). When $p>2$ (high--order) the kernels will have negative parts and are not probability densities (referred to as bias--reducing kernels).


The asymptotic efficiency of applying KDE only depends on the bandwidth $h$ and the kernel function $K(\cdot)$, and is commonly measured by the expected $L_2$ risk function, also termed mean integrated squared error ($MISE$), as defined in Eq. \ref{eq:mise}.
\begin{equation}\label{eq:mise}
MISE(h)=E\left[\int \left(\hat{f}_h(x)-f(x) \right)^2 dx\right]
\end{equation}


Scott \cite{Scott1979} proposed a data--based bandwidth for histograms ($h=3.49\sigma n^{-\frac{1}{3}}$) based on the Gaussian density, which is not a strong assumption as it will not necessarily result in shapes that looks Gaussian. Another estimate for the Gaussian kernel is $h=1.059\sigma n^{-\frac{1}{5}}$, introduced by Silverman \cite{Silverman86}, which works well for smoothed densities, and it is often called \textit{normalized reference rule}. However, if the density is multimodal, then $h$ tends to oversmooth. Silverman also suggested and adaptive estimate in the case of distributions that deviate from normality, as bimodal distributions might be overestimated by the Gaussian kernel estimate. The approach $h=0.9An^{-\frac{1}{5}}$ (where $A=min\left(\hat{\sigma},\frac{\widehat{IQR}}{1.34}\right)$, $\hat{\sigma}$ is the sample standard deviation, and $\widehat{IQR}$ is the sample interquartile range) is well suited to unimodal densities, while it does not severely overestimate bimodal distributions. 


In general, the bandwidth choice has a stronger impact than that of the form of the kernel function $K$ on the quality of the estimated density (a small value of $h$ may increase unnecessarily the variance, while a large value may cause oversmoothing, which may consequently mask some of the important features, e.g., multimodality). A detailed discussion about the bandwidth selection can be found in \cite{Turlach1993}.

In summary, KDE is a non--parametric way to estimate the \textit{pdf} of a random variable, and it is particularly useful when the underlying distribution of the data is unknown or does not conform to standard distributions like the Gaussian, allowing for a more flexible and data--driven approach to modeling the distribution of each feature.

\subsection{KDE--based naïve--Bayes}

John and Langley's work \cite{John1995} focused on enhancing the NB classifier, but their specific contribution involved using KDE as an alternative to assuming Gaussian distributions for continuous variables. By employing KDE, they proposed a method, named \textit{Flexible Naïve--Bayes} (FNB), that did not assume a specific parametric form for the distribution of the data. This flexibility allows the model to adapt to the actual shape of the data distribution, which can vary widely across different features and datasets, thus potentially improving classification performance, especially in cases where the data distributions are far from Gaussian. Therefore, FNB should perform well in domains that violate the normality assumption.


In the FNB approach, to compute the probability $P(x_j \vert c_i)$ in Eq. \ref{eq:bayes_decision}, the Eq. \ref{eq:kernel} is used, but limiting the samples to those belonging to class $c_i$, as shown in Eq. \ref{eq:kde_probability}.
\begin{equation}\label{eq:kde_probability}
P(x_j \vert c_i)\approx \hat{f}_h(x\vert c_i)=\frac{1}{n_i h}\sum_{x_j \in c_i}K\left(\frac{x-x_j}{h}\right)
\end{equation}
\noindent where $n_i$ is the total number of samples $x_j$ that belongs to class $c_i$.

The experiments carried out in \cite{John1995} only used eleven datasets with a number of variables ranging from 8 to 24, but showed that FNB performed significantly better than NB for only five datasets. However, FNB revealed worse performance with datasets containing greater number of continuous features, which might be an issue in the context of genomic datasets.

It is well known that as the dimensionality of the feature space increases, the classification performance can deteriorate, because data points become increasingly sparse, leading to degraded performance of the classifier. This phenomenon, named \textit{the curse of dimensionality}, coined by Bellman \cite{Bellman1961}, introduces more difficulty in Bayes--based approaches. For very high--dimensional spaces, as it is the case of genomic datasets, NB--based methods need to compute the product of very many probabilities, which negatively impact the model's performance as they quickly tend to zero. A strategy to mitigate this issue is to first employ feature selection or dimensionality reduction techniques (such as Principal Component Analysis) \cite{Clarke2008}. However, none of these methods consider the potential relevance of each individual class in the further classification performance. This important aspect will be considered in detail in Section \ref{sec:XNB}.

In summary, integrating KDE into the NB classifier represents an important step toward creating more adaptable and accurate probabilistic models, especially for handling continuous data in various applications. This approach demonstrates the potential for enhancing traditional models with non--parametric methods to better reflect the complexities of real--world data. 


\section{Explainable Class--Specific Naïve--Bayes}\label{sec:XNB}

\subsection{Class--specific versus class--independent}

The genomics boom has led to the emergence of datasets with thousands of variables. Furthermore, transposable elements have further increased dimensionality to the order of millions of variables \cite{Wellinger2022}. This new scenario has introduced fresh challenges for existing algorithms and has spurred research into the development of efficient techniques capable of handling such a vast number of features from two perspectives: performance and explainability. 

Feature selection techniques can be broadly categorized into two groups: class--independent and class--specific. The scientific literature on class--independent feature selection has been extensively prolific, amassing thousands of articles indexed in Scopus. However, this level of productivity has not been mirrored in the domain of class--specific feature selection, where fewer than two hundred articles have been published in journals since its inception in 1997 to date.

Class--independent techniques encompass the majority of traditional feature selection methods, such as Information Gain  \cite{Quinlan1986}, $\chi^2$ Test \cite{Plackett1983}, and ReliefF \cite{Kononenko1997}. These techniques aim to identify a global set of attributes relevant for the entire dataset without differentiating between its classes, and they are primarily used in classification tasks.

In contrast, class--specific feature selection focuses on each class independently to ascertain a relevant subset of attributes for each class. This approach allows for a tailored analysis of each class, identifying attributes that are particularly significant for each one. The results from this process can either be utilized separately for each class or combined using an aggregation method to form a comprehensive attribute subset that encompasses insights from all classes. 

The scenario becomes more complex in multi--class problems, where datasets feature more than two class labels. In these cases, a variable might be highly relevant for one class but not for others. Therefore, it is crucial that the variables selected for such datasets collectively possess the ability to discriminate each class effectively. This method is especially valuable in fields such as healthcare, where the significance of different attributes can vary greatly in identifying specific ailments or diseases.

The early stages of machine learning were dominated by the tendency to encompass all variables, which was far from the principle of feature parsimony, crucial for predictive accuracy. Nowadays, where very high dimensional contexts exist \cite{Feltes2019}, restricting the number of input features helps mitigate overfitting and enhances the models' ability to make accurate estimates. However, although there have been many efforts to reduce the dimensionality faced by a classifier, few instead have attempted to build the classifier based on a non--homogeneous set of features, i.e., feature sets selected separately for each class, as opposed to the universal approach that performs the selection jointly for the entire dataset. 

Baggenstoss made significant contributions to the development of a class--specific model of behavior, focusing on the separation of attributes by class. In his earlier works \cite{Baggenstoss1998, Baggenstoss1999}, he introduced the concept of  \textit{class--specific} modeling, aimed at estimating low--dimensional probability density functions while maintaining the theoretical effectiveness of classification. Subsequent research by Baggenstoss \cite{Baggenstoss2003,Tang2016,Baggenstoss2022} further elaborated the class--specific approach by employing an invertible and differentiable multidimensional transformation to generate new features. This approach represents a feature extraction procedure rather than a mere selection of existing attributes from the dataset. In this context, the generated features are class--specifically derived or extracted, signifying that they are not subsets of the original dataset attributes, but are instead new features created through transformation processes.

A general framework for class--specific feature selection, based on the one--against--all strategy is proposed in \cite{Pineda2011}. They transform the $L$--class problem into $L$ binary problems, one for each class, and then the classification is performed by an ensemble strategy. For each class, after one--against--all class binarization, an oversampling technique is applied to balance the dataset, and then a feature selection method extracts the relevant features for that class. Although the approach is very interesting, it focuses on building several classifiers, each using a different subset of features, and then integrating them into an ensemble scheme, which will aggregate the classification outputs. In contrast, the novelty of our approach lies in integrating the class--specific feature selection into a single classifier.

\subsection{Method}

The method presented in this work, named  \textit{Explainable Naive Bayes} (XNB), incorporates the concept of class--specific feature selection. Unlike traditional approaches that construct separate classifiers for each class, XNB integrates the selected features directly into the probability calculations of the FNB model. Specifically, XNB calculates the probabilities of variables using KDE, but it selectively includes only those features that are relevant for each class. This approach streamlines the process by focusing on pertinent features, thereby enhancing the model's efficiency and interpretability.
\begin{algorithm}
\KwData{$D$: dataset \\
$bw\_f$ : \{Scott, Silverman,$\dots$\} (bandwidth function)\\
$k\_f$ : \{Gaussian, Epanechnikov,$\dots$\} (kernel function)\\
$\mu$: 50 (number of evaluation points for KDE)\\
$\theta$: 0.999 (threshold)
}
\KwResult{$\Psi$: model}
\caption{Explainable Naïve--Bayes (XNB)}
\label{alg:XNB}
$[(c,v,bw)] \leftarrow$ Calculate\_Bandwidth$(D, bw\_f)$ \\
$[(v,c,kde)] \leftarrow$ Build\_KDE$(D, [(c,v,bw)], k\_f, \mu)$ \\
$[(v,c_i,c_j,h)] \leftarrow$ Calculate\_Hellinger$(D, [(v,c,kde)])$ \\
$[(c,[v])]  \leftarrow$ Select\_Class-Specific$(D, [(v,c_i,c_j,h)],\theta)$ \\
$\Psi=[(c,[(v,kde)])]  \leftarrow$ Build\_Model$(D,[(c,[v])])$ \\
\end{algorithm}

The function Calculate\_Bandwidth provides all the bandwidths for each pair $(class, variable)$, taking into account the algorithm $bw\_f$ to calculate the bandwidths (in this work, Silverman's rule \cite{Silverman86}). 

The function Build\_KDE builds the KDE models for each pair $(class, variable)$, considering the kernel function indicated in $k\_f$ (in this work, Gaussian), and a number $\mu$ of $x$ points for the support of the kernel (in this work, 50 has proven to be precise enough). 

The function Calculate\_Hellinger quantifies the Hellinger distance \cite{Hellinger1907,Hellinger1909} between every two probability distributions (one for each class pair, $c_i$ and $c_j$) for each variable $v$. Let $P=[p_1, \ldots, p_K]$ and $Q=[q_1, \ldots, q_K]$ be two discrete probability distributions. The Hellinger distance is defined as: 
\begin{equation}
H(P,Q)=\frac{1}{\sqrt{2}} \sqrt{ \sum_{j=1}^K \left( \sqrt{p_j}-\sqrt{q_j} \right)^2}  \label{eqn:Hellinger}
\end{equation}

The choice of the Hellinger distance is due to its interesting properties: a) it is symmetric; b) bounded (defined in [0,1]); c) convex with respect to both $P$ and $Q$; and d) it satisfies the triangle inequality. As a consequence, it is recently attracting considerable amount of attention in many scientific fields \cite{Zhao2017,Cartolano2020,Peng2020,Wasserman2020,Zhu2021,Rochman2021,Batzianoulis2021,Rosswog2021,Duren2021, Aguilar2022a, Aguilar2024a}. From our perspective, if two distributions ($c_i$ and $c_j$) are very different ($H(c_i,c_j))\!\approx\!1$) for the same variable $v$, then $v$ might have high discriminatory power for $c_i$ and $c_j$.

The function Select\_Class-Specific is designed to determine the smallest subset of relevant variables $[v]_{c_i} $ that can effectively discriminate a specific class $c_i$ from the other classes. This selection is based on a threshold parameter $\theta$, which plays a crucial role in selecting how many variables are sufficient to achieve the desired level of discrimination. In Eq.  \ref{eq:criterion}, $[v]_{c_i}$ represents the smallest subset $S$ of features from the set $F$ that ensures the discriminatory power  $D(S, c_i)$ exceeds the threshold $\theta$. The  $\min$ operator indicates that the function seeks the subset with the minimum cardinality.

\begin{equation}\label{eq:criterion}
[v]_{c_i} = \min \left\{ S \subseteq F \left\vert  D(S,c_i) > \theta \right. \right\}
\end{equation}

The discriminatory power  $D(S, c_i)$, defined in Eq. \ref{eq:discriminatory_power}, quantifies the ability of the subset $S$ to distinguish the class $c_i$ from the other classes $c_j$ based on the Hellinger distance $H\left(c_i,c_j\vert v\right)$ for each feature $v \in S$. The product term accounts for the cumulative effect of all features in $S$, and subtracting this product from 1 measures the overall discriminatory strength.

\begin{equation}\label{eq:discriminatory_power}
D(S,c_i) = 1-\prod_{\substack{c_j\in \mathbb{L}\\ v\in S}} \left(1-H\left(c_i,c_j\vert v\right) \right)
\end{equation}

\begin{figure*}[t]
\centering
\includegraphics[width=1\linewidth]{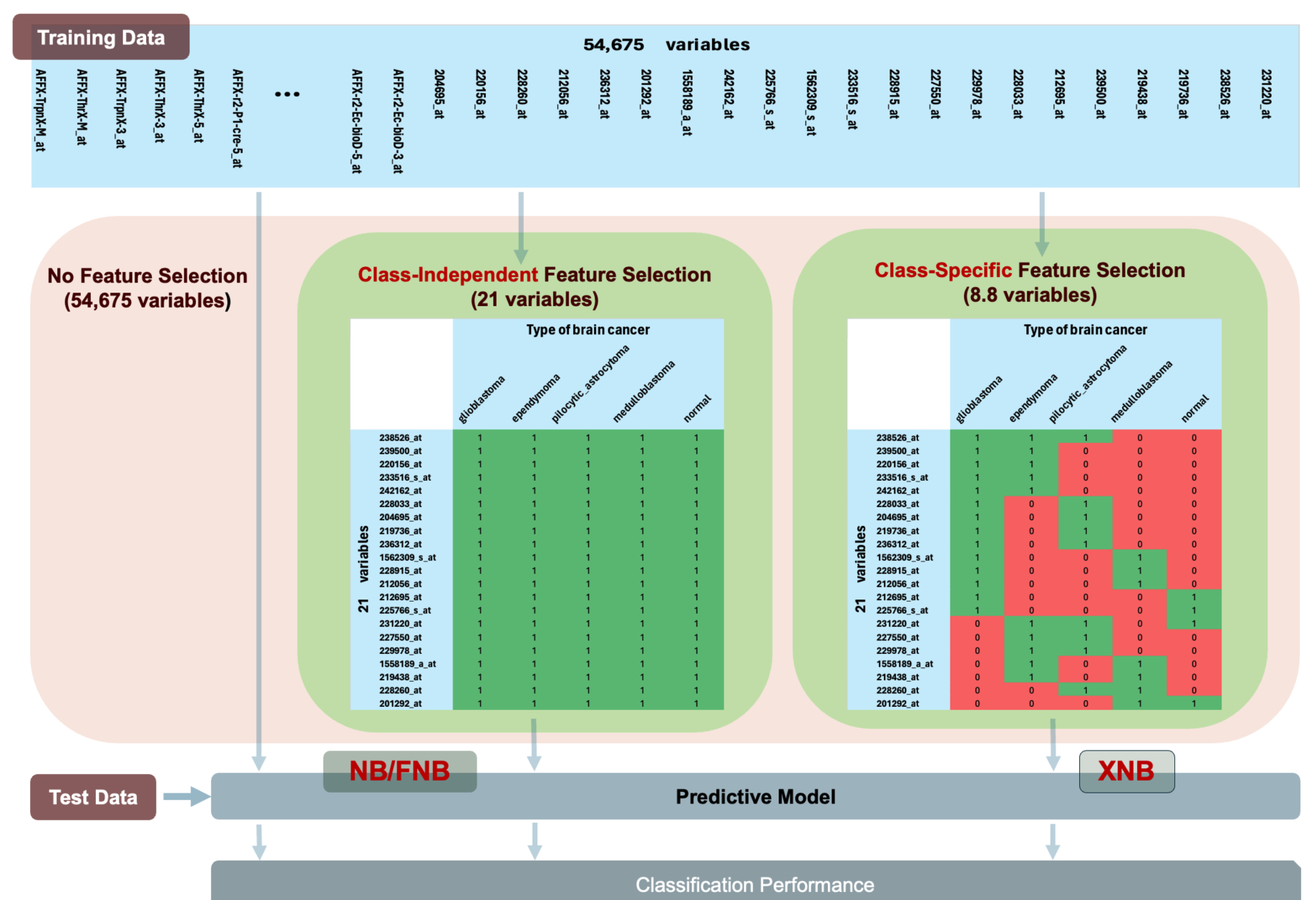}
\caption{Comparative scheme of the methodology. Example of Brain GSE50161, with 54,675 variables and 5 classes. It is shown the result for one fold from stratified 10--fold cross--validation. Total number of unique variables identified after feature selection was 21 (class--independent approach). The number of variables for each class was \#vars = [14, 10, 9, 7, 4], with a mean of 8.8 (class--specific approach). Accuracy for NB was 0.846 (using all the original variables), and for XNB was 1.000.}
\label{fig:scheme}
\end{figure*}

When the greatest Hellinger distance, $H(c_i,c_j\vert v)$, is greater than $\theta$, that means that only one variable $v$ is able to discriminate both classes $c_i$ and $c_j$, from each other. When $H(c_i,c_j\vert v)\leq \theta$, then it is necessary to search for another variable $v'$, with high $H(c_i,c_j\vert v')$, such that both $v$ and $v'$ can contribute to exceed $\theta$. The threshold $\theta$ is set to 0.999, which ensures that only the most discriminative features are selected for each class. The process continues until the subset $S$ of variables exceeds $\theta$ or all the variables have been selected, i.e., $S=F$. However, experimental results show that it is very unlikely to build $S$ with all the original variables, but rather $S$ usually have small cardinality. In case of genomic datasets the reduction is extraordinary, containing $S$ less than 0.02\% of $\vert F \vert$ on average, and maintaining the classification performance. Minimizing the cardinality of $S$ reduces the dimensionality of the model, leading to more interpretable and efficient classifiers.

The function Build\_Model builds the KDE models for all the classes, only considering the variables selected in the previous step. It is interesting to note that the FNB approach would omit the steps Calculate\_Hellinger and Select\_Class-Specific, and would build the model $\Psi$ with all the variables for each class. On the contrary, the NB classifier would only use Build\_Model to calculate the Gaussian estimators (without KDE), and would also use all the variables for each class. 

Figure \ref{fig:scheme} (green square on right side) shows an example of output from function Select\_Class-Specific, in which a list $[(c,[v])]$ is returned. From each class $c$ (glioblastoma, ependymoma, pilocytic astrocytoma, medulloblastoma, normal) in the dataset Brain\_GSE50161, with 54,675 variables, a list of associated variables $[v]$ is shown (1 means present, and 0 not present). For instance, only the variables \{1562309\_s\_at,  228915\_at, 212056\_at, 1558189\_a\_at, 219438\_at, 228260\_at, 201292\_at\} are associated to class medulloblastoma. Any traditional class--independent feature selection technique would have produced a table entirely filled with ones (green square on the left side) across all variables selected by that technique, rather than only including the 21 variables obtained by combining those involved in the class--specific result on the right.


The prediction is then made by adapting Eq. \ref{eq:bayes_decision} to the class--specific concept, i.e., by following Eq. \ref{eq:x_bayes_decision}, in which not all the variables are considered for each class $c_i$, but only those that are present in the model $\Psi$ for each $c_i$, which are likely different, as depicted in Figure \ref{fig:scheme}.
\begin{equation}\label{eq:x_bayes_decision}   
c^*=\argmax\limits_{i \in \{1, \ldots, k\}} P(c_i) \displaystyle\prod_{v_j \in \Psi(c_i)} P(v_j \vert c_i)
\end{equation}

To calculate all probabilities $P(v_j \vert c_i)$, the KDE models associated to class $c_i$ and variable $v_j$, $\Psi(c_i,v_j)$, provided by Alg. \ref{alg:XNB}, are used.

\subsection{Complexity}

The complexity of calculating the bandwidths and building the KDE models is $O(mnk)$, where $m$ is the number of variables, $n$ is the number of instances, and $k$ is the number of classes. Determining the Hellinger distances between each pair of KDE models for each variable and selecting the relevant variables for each class has a complexity of $O(mnk^2)$. Subsequently, the complexity of building the final model is $O(mnk)$. Thus, the overall complexity of XNB is $O(mnk^2)$. This represents a slight increase by a factor of $k$ compared to the complexity of NB, which is $O(mnk)$. However, this increase is not significant as the number of classes is typically very low compared to the number of variables or samples ($k \ll m$ and $k \ll n$). The small added complexity is a worthwhile trade--off for the increased explainability and potential improvements in classification performance, particularly in high--dimensional spaces.

\section{Experiments}\label{sec:experiments}

A comparative analysis was performed with datasets chosen from the Curated Microarray Database (CuMiDa)\cite{Feltes2019}, with varying number of samples (21--281), very high number of variables ranging from about 22,277 to 54,675, and number of classes from 2 to 7. Detailed description of datasets is shown in Table \ref{tab:datasets}. The total number of real values processed ranges from 1.04 millions (GSE11682) up to 8.26 millions (GSE45827).

All the classification performance results were obtained by averaging the results of stratified 10--fold cross--validation. For each dataset two classification methods were applied: NB (the standard Naïve--Bayes), and XNB (eXplainable class--specific Naïve--Bayes). Table \ref{tab:results} shows that classification performance for both methods is very similar (see last row: Mean). However, there is an outstanding improvement in terms of the number of factors involved in the model, i.e., of the number of explanatory variables in the process of cancer prediction. Last row in Table  \ref{tab:results} shows also the mean of number of variables involved in the classification model for XNB, achieving an extremely low value of 8.3, which means a remarkable average reduction of the feature space dimensionality of about 99.98\%.

\begin{table}[t]
\caption{Results for CuMiDa datasets. First two columns are described in Table \ref{tab:datasets}. NB Acc and XNB Acc refer to the accuracy of NB and XNB, respectively. Last column, XNB \#v, shows the mean of variables selected by XNB. Last row (Mean) stands for the mean of NB Acc, XNB Acc, and XNB \#v, respectively.}
\label{tab:results}
\begin{center}
\begin{tabular}{|l|c||r|r|r|}
\hline
Type 		& GSE 	& NB Acc		& XNB Acc 	& XNB \#v	\\ \hline \hline
Bladder		& 31189	& 0.507		& 0.564		& 12.6		\\ \hline
Brain			& 15824	& 0.700		& 0.575		& 5.7			\\ \hline
Brain			& 50161	& 0.915		& 0.915		& 10.7		\\ \hline
Breast		& 07904	& 0.940		& 0.980 		& 3.1			\\ \hline
Breast		& 10797	& 0.683		& 0.605		& 10.4		\\ \hline
Breast		& 26304	& 0.418		& 0.444 		& 35.5		\\ \hline
Breast		& 42568	& 0.922		& 0.992		& 2.8			\\ \hline
Breast		& 45827	& 0.933		& 0.887 		& 7.9			\\ \hline
Colorectal		& 21510	& 0.993		& 0.980 		& 3.5			\\ \hline
Colorectal		& 77953	& 0.837		& 0.810		& 6.8			\\ \hline
Leukemia 		& 28497 	& 0.822 		& 0.829 		& 13.8 		\\ \hline
Lung			& 07670	& 0.927		& 0.923		& 3.5			\\ \hline
Ovary		& 06008	& 0.725		& 0.684		& 10.9		\\ \hline
Pancreatic	& 16515	& 0.823		& 0.900		& 4.1			\\ \hline
Prostate		& 11682	& 0.525		& 0.617		& 6.9			\\ \hline
Prostate		& 46602	& 0.920		& 0.940		& 2.9			\\ \hline
Renal		& 53757	& 0.839		& 0.852		& 6.1			\\ \hline
Throat		& 59102	& 0.955		& 0.950		& 2.8			\\ \hline \hline
Mean		&	--	& 0.799		& 0.803		& 8.3			\\ \hline
\end{tabular}
\end{center}
\end{table}


Figure \ref{fig:scheme} illustrates that the average number of variables needed for the classification task on the Brain GSE50161 dataset, with 54,675 variables and 5 classes, is 8.8. The XNB model also provides detailed information on the specific variables involved in the prediction of each type of cancer. While the type glioblastoma needs 14 variables, the type medulloblastoma only needs 7 variables. In addition, discriminating normal from cancer is possible with only 4 variables. 


The explanatory and interpretable potential of the solutions provided by XNB is highly valuable. XNB not only achieves good classification performance but also establishes associations between classes and variables with high discriminatory power. 


\section{Conclusions}\label{sec:conclusions}


Class-specific strategies concentrate on evaluating the discriminatory power of features within each class individually, which enables a more refined understanding of how features correlate with each specific class. This method can lead to enhanced discrimination between classes, as it allows for identifying the most informative features for each class separately. As a result, the interpretability of the classification model is improved, and it facilitates the extraction of domain--specific insights. Additionally, class--specific strategies are inherently more resilient to class overlap, where classes may share common feature distributions, by focusing on the distinctiveness of features within each class.


Explainability plays a critical role in understanding how features contribute to predictions for different classes. Class--specific strategies enhance this explainability by allowing for the analysis of why particular features are deemed important or discriminatory for specific classes. This understanding is essential for fields like personalized medicine, where knowing the rationale behind model predictions can inform clinical decisions.

While the naïve--Bayes assumption assumes that features are conditionally independent given the class, this assumption may not hold for all classes. Analyzing variables from a class--specific perspective helps identify features that are not independent for certain classes because they share similar probability distributions, and will be discarded as discriminatory variables for those classes. 


As shown in the experimental analysis the number of variables involved in the classification model has been drastically reduced (using on average less than 0.02\% of the original number of variables).  
Despite this important reduction, the model maintains high classification performance, with an average of about 8 variables per dataset. This reduction makes the model significantly more explainable, reducing the risk of overfitting while maintaining the quality of the predictive model.


In summary, XNB significantly enhances the explainability of the model by reducing the feature space to a minimal subset for each class. This not only improves interpretability but also reduces the risk of overfitting, making the model more robust. The potential applications of XNB in real--world scenarios, particularly in high--dimensional domains, where both explainability and accuracy are crucial, are substantial and promising.



Several aspects would require more analysis for improving the XNB approach in future work. For instance, when the underlying density has long tails or in presence of multimodality, adaptive bandwidth methods might offer better fitting by regulating the trade--off between variance and bias. Although the performance of the Hellinger distance is excellent, determining which is the most informative distance function would require an exhaustive analysis of not only distance metrics but also divergence measures for probability distributions (Jeffreys distance \cite{Jeffreys1946}, Wasserstein distance \cite{Kantorovich1939} or  Kullback--Leibler divergence \cite{Kullback1951}, among others).





\section*{Software availability}
  
The XNB source code is available in Github at \url{https://github.com/sorul/xnb}.

\section*{Acknowledgment}

This work was supported by Grants PID2020-117759GB-I00 and PID2023-152660NB-I00 funded by the Ministry of Science, Innovation and Universities.

\bibliographystyle{unsrt}  
\bibliography{bibliography}



\end{document}